\def\year{2022}\relax
\documentclass[letterpaper]{article} 
\usepackage{aaai22}  
\usepackage{times}  
\usepackage{helvet}  
\usepackage{courier}  
\usepackage{nameref}  
\usepackage[hyphens]{url}  
\usepackage{graphicx} 
\usepackage{mathtools}
\usepackage{natbib}  
\usepackage{caption} 
\DeclareCaptionStyle{ruled}{labelfont=normalfont,labelsep=colon,strut=off} 
\usepackage{algorithm}
\usepackage{algorithmic}

\usepackage{newfloat}
\usepackage{listings}
\floatstyle{ruled}
\newfloat{listing}{tb}{lst}{}
\floatname{listing}{Listing}
\usepackage{booktabs}
\usepackage{tikz}
\usetikzlibrary{arrows,shapes, decorations.pathmorphing,backgrounds,positioning}
\tikzset{every picture/.style={/utils/exec={\sffamily}}}
\usepackage{subcaption}
\usepackage{amsmath}
\usepackage{multirow}

\title{Modeling Multi-level Context for Informational Bias Detection by Contrastive Learning and Sentential Graph Network}
\author {
    Shijia Guo,\textsuperscript{\rm 1}
    Kenny Q. Zhu, \textsuperscript{\rm 1}
}
\affiliations {
    \textsuperscript{\rm 1} Shanghai Jiao Tong University\\
    seveny1997@sjtu.edu.cn, kzhu@cs.sjtu.edu.cn
}

\begin{document}

\maketitle

\begin{abstract}
Informational bias is widely present in news articles. It refers to providing one-sided, selective or suggestive information of specific aspects of certain entity to guide a specific interpretation, thereby biasing the reader's opinion. Sentence-level informational bias detection is a very challenging task in a way that such bias can only be revealed together with the context, examples include collecting information from various sources or analyzing the entire article in combination with the background. In this paper, we integrate three levels of context to detect the sentence-level informational bias in English news articles: adjacent sentences, whole article, and articles from other news outlets describing the same event. Our model, MultiCTX (Multi-level ConTeXt), uses contrastive learning and sentence graphs together with Graph Attention Network (GAT) to encode these three degrees of context at different stages by tactically composing contrastive triplets and constructing sentence graphs within events. Our experiments proved that contrastive learning together with sentence graphs effectively incorporates context in varying degrees and significantly outperforms the current SOTA model sentence-wise in informational bias detection.

\end{abstract}

\section{Introduction}
Informational bias broadly exists in news articles. As a sort of framing bias, it always frames a certain entity by specific aspects using narrow, speculative or indicative information  to guide a particular interpretation, thus swaying readers' opinion. 

For most of us, news articles are the main source of information. Therefore, news articles play a central role in shaping individual and public opinions. However, news reports often show internal bias. The current research is often limited to the lexical bias. This form of bias rarely depends on the context of the sentence. It can be eliminated by deleting or replacing a small number of biased words. Contrarily, researchers \citet{fan-etal-2019-plain} found that the informational bias is more common and more difficult to detect.

Different from other types of bias, the sentence-level informational bias detection largely depends on the context and this fact makes the task very challenging. A sentence alone can be expressed in a neutral manner, but it might be revealed as biased in consideration of the context. Take the second row in Table \ref{tab:basil} as example: the sentence \textit{``Mr. Mattis, a retired four-star Marine general, was rebuffed.''} \label{sent:mattis} seems to be a very simple declarative sentence stating a fact. However, if we read the previous sentence \textit{``Officials said Mr. Mattis went to the White House with his resignation letter already written, but nonetheless made a last attempt at persuading the president to reverse his decision about Syria, which Mr. Trump announced on Wednesday over the objections of his senior advisers.''} (the first row in Table \ref{tab:basil}) , we will know that \textit{`a retired four-star Marine general'} indicates a negative, even ironic tone towards Mr. Mattis and his last attempt. Therefore, sentence-level informational bias can only be revealed by collecting information from various sources and analyzing the entire article together with its background. Such subtleties of informational bias are more likely to affect unsuspecting readers, which indicates the necessity of research into new detection methods. 

In this paper, we propose MultiCTX (Multi-level ConTeXt), a model composed of contrastive learning and sentence graph attention networks to encode three different levels of context:  \textbf{1) Neighborhood-context}: adjacent sentences, i.e. sentences in the same article around the target sentence; \textbf{2) Article-context}: the whole article containing the target sentence; \textbf{3) Event-context}: articles from various news media reporting the same event. These three levels encompass the contextual information from the most local to the most global.

In order to make use of the context rather than be overwhelmed by the noise introduced, MultiCTX prioritizes contrastive learning which learns sentence embedings via discriminating among $(target, positive$ $sample, negative$ $sample)$ triplets to distill the essence of the target sentence. The quality of the learned CSE (Contrastive Sentence Embedding) relies on that of triplets. Other than the traditional brute-force way to select triplets only based on their labels, MultiCTX further considers article-level information which creates higher-quality triplets. Such triplet formulation guarantees that our CSEs infuse the context and reflect sentences' inherent semantics instead of the shallow lexical features.

MultiCTX then builds a relational sentence graph using CSEs. Edges are connected between two sentences if they are logically related in the same \textit{neighborhood} or if they are continuous in entities or semantically similar within the same \textit{event}. Finally we apply a Self-supervised Graph Attention Network (SSGAT) on our sentence graph to make the final informational bias prediction. The SSGAT structure encodes neighborhood-level and event-level context via edges, making it possible for textually distant but contextually close sentences to connect directly. The flexible graph structure extends beyond the sequential arrangement of traditional LSTMs, which also consider the surrounding context. 

Although document graphs are not rare in NLP tasks, they are often short and built by token-wise dependency parsing. It may suffer from high complexity and considerable noise when applied on long texts which is our case with news articles. Our relational sentence graph uses sentence nodes and focuses on inter-sentence relationships. It recquires only minimal syntax parsing, takes on less noise and has better interpretability.

Few research studies sentence-level informational bias detection by infusing context. \citet{fan-etal-2019-plain} first published a human-annotated dataset on this task, taking the context into account during annotation. However, sentences are still treated sentences individually in their model. \citet{van-den-berg-markert-2020-context} did a primary research on incorporating different levels of context in the informational bias detection. However, they consider only one kind of context in each model. To our best knowledge, our model is the first to incorporate multi-level contextual information in sentence-level classification task.

In summary, we present the following contributions:

\begin{itemize}
  \item We are the first to incorporate three different levels of context together in the sentence-level bias detection task. By introducing context, we aim to simulate how people learn new things in real life: widely reading, generally picturing and thoroughly reasoning.
  \item We propose a novel triplet formation for contrastive learning in bias detection. The methodology can be generalized for other tasks.
  \item We are the first to use a sentence graph to encode the textual context information in the bias detection task.
  \item Our model MultiCTX significantly outperforms the current state-of-the-art model by 2 percentage points F1 score. It indicates that contextual information effectively helps sentence-level informational bias detection and our model successfully infuses multi-level context.
\end{itemize}

\section{Methodology}

\begin{figure*}[!htbp]\centering
\includegraphics[width=\linewidth]{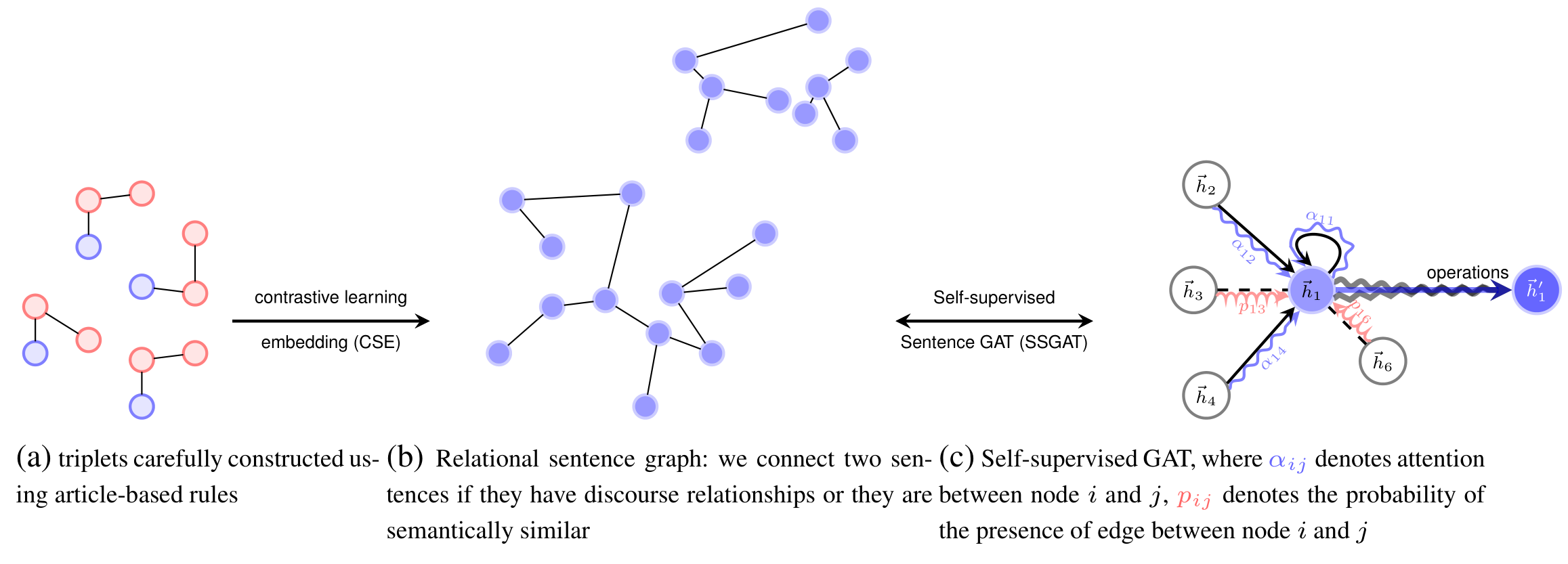}
  \caption{ Our Model MultiCTX }
  \label{fig:model}
\end{figure*}

Figure \ref{fig:model} illustrates our model MultiCTX (Multi-level ConTeXt). First, we carefully construct triplets from the original dataset and then apply supervised contrastive learning on them to obtain sentence embeddings. Second, we build relational sentence graphs by joining sentence nodes according to their discourse relationships and semantic similarity. Finally, we apply a Self-supervised Graph Attention Network \citep{kim2021how} to perform the bias detection as a node classification task. In essence, MultiCTX has two modules, Contrastive Learning Embedding (CSE) and Self-supervised Sentence Graph Attention Network (SSGAT). In order to investigate the role of the context and to imitate the way people learn from the news reports, we also apply a more reasonable and challenging cross-event data splitting. 

\subsection{Data splitting}

First of all, let's think about the nature of the news reports and the way people learn about the world in real life. News articles always emerge almost simultaneously in large numbers along with a particular event, over which people reason based on their experience learnt from previous events. Moreover, people usually read an article as a whole instead of randomly picking up several sentences and they are unlikely to encounter a sentence from news events happened before. Additionally, people tend to collect information from more than one article to get a bigger picture of the new event. Therefore, in order to simulate the real human's learning process, different from the commonly-used data splitting which randomly distributes sentences to one of the three subsets (train/val/test), we use event-wise data splitting mentioned in \citet{van-den-berg-markert-2020-context}, \citet{chen-etal-2020-detecting}. We treat the articles reporting the same event as a unit and keeping sentences from the same event in the same subset. Part of the data is shown in Figure \ref{tab:basil} with clear `adjacent sentences', `article' and `event' structure.

Furthermore, splitting by events is more reasonable and more demanding, in terms of model generalizability for identifying informational bias from unseen events. Experiments in \citet{van-den-berg-markert-2020-context} and \citet{chen-etal-2020-detecting} also show that common models including BERT-based models all experienced a considerable performance drop when switching from random splitting to event-based splitting.

\begin{table*}[th]
    \centering
    \begin{tabular}{cccp{12cm}c}
        \toprule
        \textbf{Event} & \textbf{Source} & \textbf{Index} & \textbf{Sentence} & \textbf{Label} \\
        \midrule
        86 & nyt & 3 &  Officials said Mr. Mattis went to the White House with his resignation letter already written, but nonetheless made a last attempt at persuading the president to reverse his decision about Syria, which Mr. Trump announced on Wednesday over the objections of his senior advisers. & 0\\
        \hline
        86 & nyt & 4 &  Mr. Mattis, a retired four-star Marine general, was rebuffed. & 1\\
        \hline
        86 & nyt & 5 &  Returning to the Pentagon, he asked aides to print out 50 copies of his resignation letter and distribute them around the building. & 0 \\
        \midrule
        11 & fox & 20 & However, Democrats rejected the plan even before Trump announced it, and a Senate version of the plan failed to get the 60 votes needed on Thursday.  & 1\\
\hline
        11 & fox & 21 & A second bill, already passed by the Democrat-controlled House to re-open the government, also fell short. & 0\\
        \midrule
        2 & hpo & 10 & There were roughly 520,000 arrests for unauthorized border crossings last year, which is about one-third of the 1.6 million arrests that happened in 2000. & 0\\
        \hline
        2 & hpo & 11 &Since 2014, a high proportion of those crossing have been Central American children and families seeking to make humanitarian claims such as asylum. & 1\\
        \bottomrule
    \end{tabular}
    \caption{BASIL dataset}
    \label{tab:basil}
\end{table*}

\subsection{Sentence Embedding using Contrastive Learning}

The idea of contrastive learning is that humans discriminate objects by ``comparison'', thus similar objects should be close to each other in the representation space, and different objects should be as far apart as possible. However, news sentences inherently have small differences in terms of pure text. Two sentences with opposite stances might be different in a few words, while two sentences expressing the
same idea are likely to be formulated completely differently. To address the problem, we apply supervised contrastive learning with hard negatives described in \citet{gao2021simcse}. The idea is to develop, from the original dataset, the triplets $(x_i,x^+_i,x^-_i)$  each denotes target sentence, positive sample and negative sample respectively. Using the $\mathbf{h}_{i},\mathbf{h}_{i^+},\mathbf{h}_{i}^-$ representations of $x_i,x^+_i,x^-_i$, 
the objective function to minimize is InfoNCE Loss.

The difficulty is to mine the positive and negative samples for each target sentence from the original dataset. A good positive sample is supposed to capture the most essential features of the sentence, rather than being influenced by other factors, such as the writing styles of different news media. Therefore, the best positive sample is expected to be significantly different from the target sample in terms of sentence formation, while the best negative sample should be similar to the target sentence in terms of syntactic structure. In short, samples with different labels from the target sentence but with initial embedding in its vicinity are likely to be the most useful, providing significant gradient information during the training process.

\begin{figure}[!htbp]\centering
    \includegraphics[width=.7\linewidth]{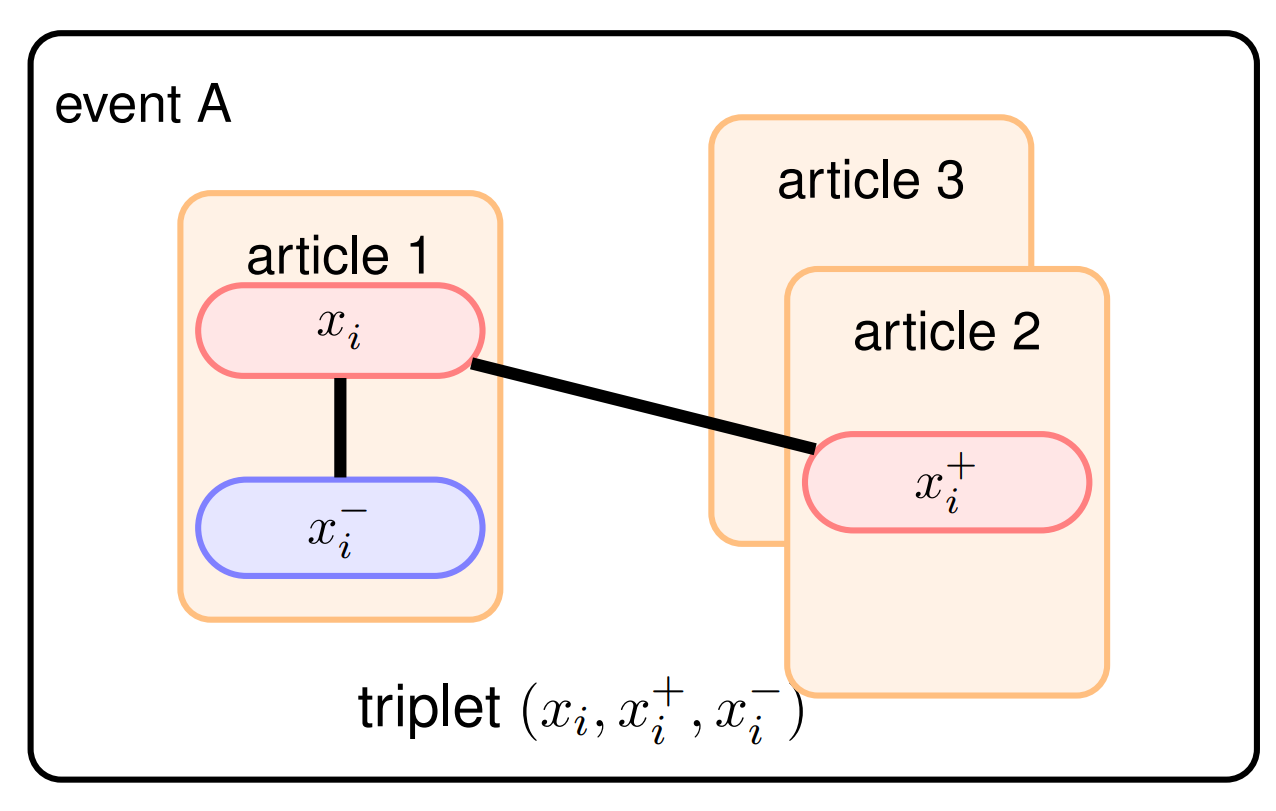}
      \caption{Triplet construction. Positive sample $x_i^+$ has same label (red) and event with target sentence $x_i$; negative sample $x_i^-$ has different label (blue) but from the same article with $x_i$; note that three sentences must from same event.}
  \label{fig:triplet}
\end{figure}

Inspired by \citet{baly-etal-2020-detect} which applies a triplet loss in training using news media in triplet selection, our final triplet follows article-based criteria and is composed of: $x_i$: target sentence; $x_i^+$: same label and event with $x_i$, but from a different article; $x_i^-$: from the same article with $x_i$ but with a different label. Figure \ref{fig:triplet} illustrates our triplet construction.

Thereby we essentially augment the original 7977-sentence corpus to a much larger dataset of around 300k triplets where sentences are no longer isolated but linked to two others. 

More importantly, triplets with the same target sentence provide altogether a microscopic `context' for the target sentence to help its representational learning. This process naturally incorporates article-level context and event-level context information:

\begin{itemize}
    \item The negative samples are from the same article as the target sentence, thus they provide an article-level context. Written by the same author, these sentences would be similar in writing styles and lexical patterns. Therefore, that article-level context made by negative samples not only informs the necessary background story information, but more importantly, they show a layer of ``skin'' of the article, forcing the contrastive learning model to uncover the superficial skin of rhetoric and wording and to truly understand the article.
    
    \item All samples are from the same event, so they provide event-level contextual information. Therefore, from the perspective of the target sentence, all positive and negative samples provide it with a small but complete world that covers most of the event information, from which the model is allowed to freely and massively draw information and to get a broad and general overview of events. Therefore the target representations can be both comprehensive and fair. Additionally, since the positive samples are from different news media and the negative samples are from the same news media as the target sentence, our model can exclude the influence of different news media writing styles and is encouraged to learn the essential meanings. In summary, our triplet construction process simulates the human learning principle of wide reading.
\end{itemize}

\subsection{Relational Sentence Graph}

\begin{figure}[!htbp]
  \centering
  \begin{subfigure}{0.8\linewidth}
    \centering
    \includegraphics[width=\linewidth]{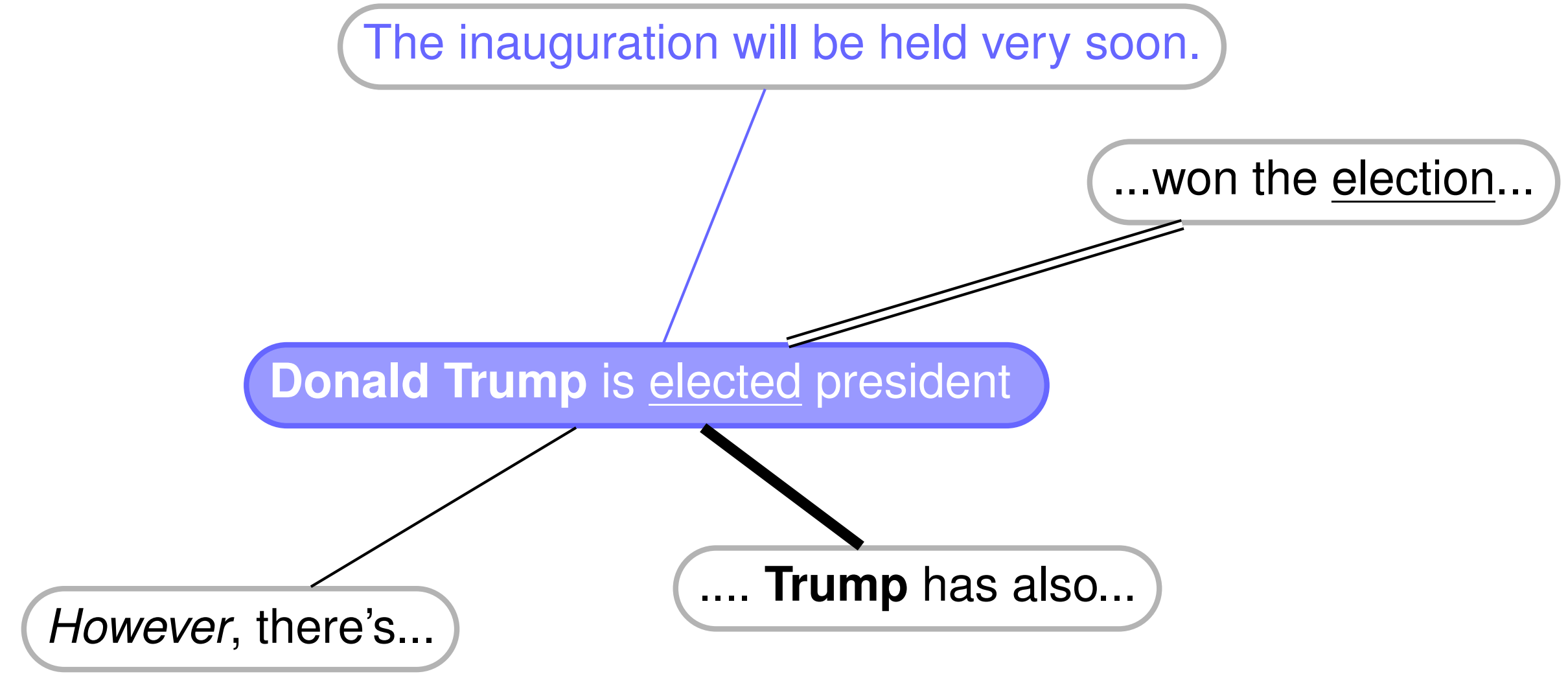}
    \caption{Four types of edges in relational sentence graph. \underline{Underline}: deverbal noun reference; \textit{italic}: discourse maker; \textbf{bold}: entity continuation; \textcolor{blue!60}{colored}: sentence similarity}
    \label{subfig:edgetype}
  \end{subfigure}
  \begin{subfigure}{0.45\linewidth}
    \centering
    \includegraphics[width=\linewidth]{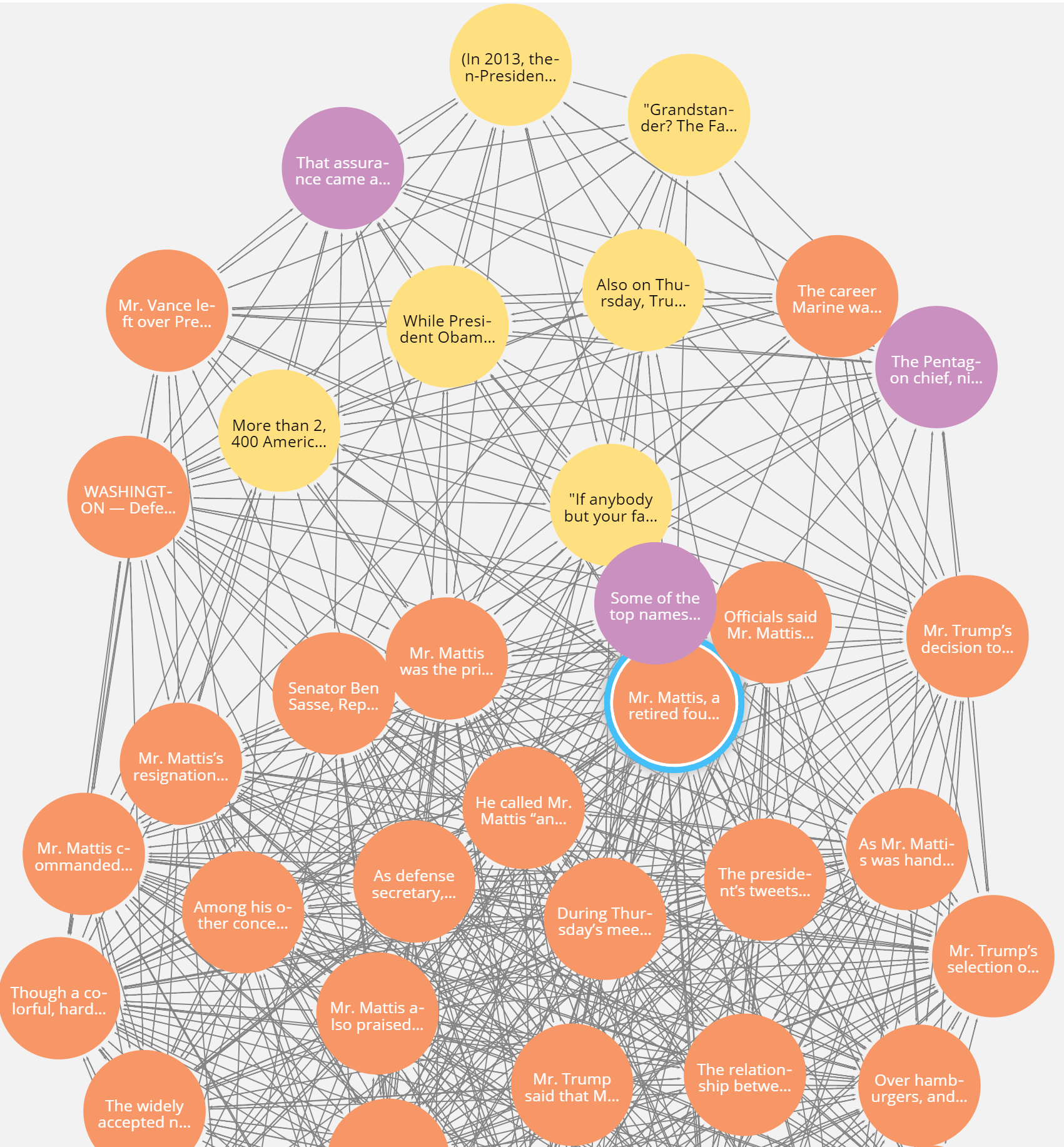}
    \caption{Partial relational sentence graph, where nodes are colored by news source: HPO (yellow), NYT (orange), FOX (purple)}
    \label{subfig:mattissrc}
  \end{subfigure}
  \begin{subfigure}{0.45\linewidth}
    \centering
    \includegraphics[width=\linewidth]{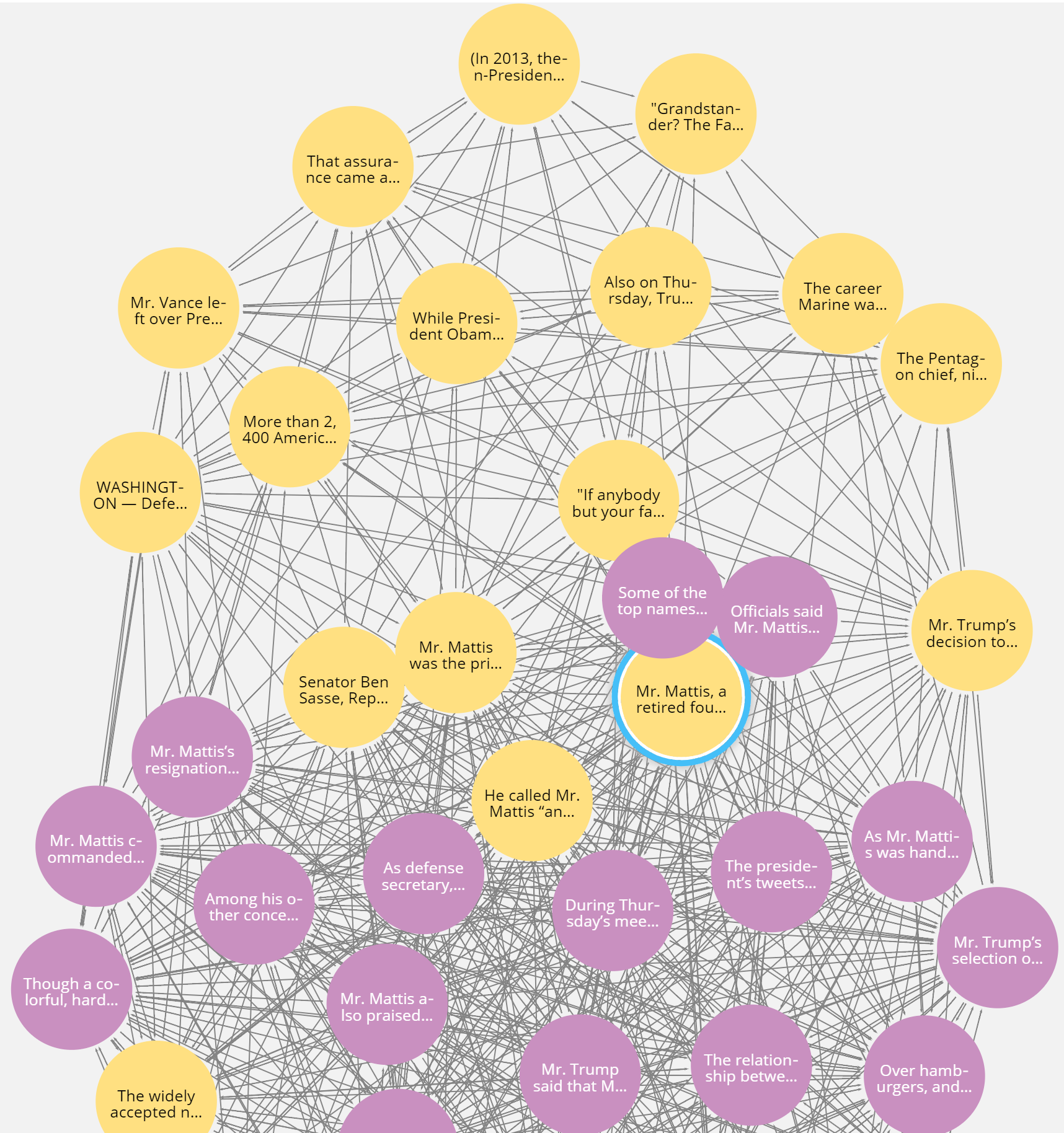}
    \caption{Partial relational sentence graph, where nodes are colored by bias label: biased (yellow), non-biased (purple)}
    \label{subfig:mattisbias}
  \end{subfigure}
   \caption{Relational sentence graph} 
  \label{fig:adg}
\end{figure}

Sentences are naturally suitable as nodes when encoding long documents, so we borrowed the idea from extractive text summarization from \citet{christensen-etal-2013-towards} and \citet{summpip} to construct graphs. The graphs are formed by connecting the sentences in four different ways illustrated in \ref{subfig:edgetype}:

\begin{enumerate} \label{para:edge}
    \item Deverbal noun reference: when an action in verb form occurs in the current sentence, it is likely to be mentioned in the noun form in the following sentences. So we attach the current sentence with its downstream sentence when at least one semantically similar deverbal noun is found in the latter.
    
    \item Discourse marker: If the immediately subsequent sentence begins with a discourse marker (e.g., however, meanwhile, furthermore), the two sentences are linked.
    
    \item Entity continuation: we connect two sentences in the same event if they contain the same entity.
    
    \item Sentence similarity: sentence pairs in the same event with high cosine similarity are joined.
    
\end{enumerate}

The four types of edge formation take different degrees of context into account: Type 1 and Type 2 consider only the subsequent sentences in the same article (neighborhood-context). In particular, Type 2 considers only the immediately following sentence. Type3 and Type 4 are not limited to adjacent sentences. Rather they consider the whole event (event-context). Note that edges occur only between in-event sentences, which is consistent with our event- based splitting.

Figure \ref{subfig:mattissrc} and \ref{subfig:mattisbias} present the same subgraph taken directly from the real relational sentence graph in our study. The two figures are only dfferent in colors, where the nodes in Figure \ref{subfig:mattissrc} are colored according to the news source, i.e. HPO (yellow), NYT (orange), FOX (purple); while Figure \ref{subfig:mattisbias} on the right is colored according to labels, i.e. biased (yellow), non-biased (purple). This subgraph presents all edges connected to the sentence \textit{``Mr. Mattis, a retired four-star Marine general, was rebuffed." (NYT)} described in Section \ref{sent:mattis}, and we see that the first sentence in Table \ref{tab:basil} is effectively linked to it. Moreover, we see that the relational sentence graph infuses information from different news media. Most sentences related to the target sentence in Figure \ref{subfig:mattisbias} from HPO (yellow) and FOX (purple) are biased (yellow) according to Figure \ref{subfig:mattisbias}. Therefore event-context contained in articles of different news outlets effectively helps identify the biased sentences.

Our graph composition is intended to mimic the way humans develop views: people acquire information through immediate context in article and reason by aggregating certain background knowledge from different news reports of the whole event.

Note that during the formation of our relational sentence graph, edges only appear between sentences in the same event.  It is compatible with our event-based data splitting, and ensures that SSGAT can be trained on the entire graph without data leakage.

\subsection{Graph Attention Network}

As one of the representative graph convolutional networks, Graph Attention Networks (GATs) introduces an attention mechanism to achieve better neighbor aggregation. By learning the weights of the neighbors, GAT can learn the representation of the target node by implementing a weighted aggregation of the neighbor node representations. However, it may suffer from graph noise introduced by incorrect node linking. In our study, we use Self-supervised Graph Attention Network \citet{kim2021how} which introduces, on top of the GAT, an edge presence prediction task and thus puts an emphasis on more on distinguishing misconnected neighbors.

The graph structure naturally places each sentence within its context, and as a result, different sentences are no longer isolated. The flexibility of the graph structure also allows it to move beyond the ordered arrangement of traditional LSTM. Therefore two sentences can be directly connected by edges, even if they are far apart in the original article or in different articles. 

Note that our sentence graph doesn't contain edges between two events, therefore it assures no data leakage while training GAT on the whole graph.

\section{Experiment and Results}

We use BASIL (Bias Annotation Spans on the
 Informational Level) dataset proposed by \citet{fan-etal-2019-plain} for the sentence-level informational bias detection task.
We experiment with four baselines including the current state-of-the-art model and four variants of MultiCTX in order to fully demonstrate each module's utility. Our results suggest that MultiCTX greatly outperforms the current SOTA and effectively incorporates the contextual information in sentence-level informational bias detection.

\subsection{Data}

 BASIL dataset provides sentence-by-sentence span-level annotation of informational bias for 300 English news articles grouped in 100 triplets, each discussing on the same event from three news outlets. The articles are selected in order to make a fair coverage in terms of time and ideology: 1) From 2010 to 2019, 10 events are included each year in the dataset; 2) Fox News (FOX), New York Times (NYT) and Huffington Post (HPO), representative of conservative, neutral and liberal respectively in the US journalism, are chosen as three news sources.

As for the sentence-level informational bias detection task, we use the same data formulation in \citet{van-den-berg-markert-2020-context}. In this sentence-wise binary classification task, a sentence is labeled as biased if at least one informational bias span occurs, and seven empty sentences are removed, resulting in a total of 7977 sentences with 1221 annotated bias.

Examples are shown in Table \ref{tab:basil}.

\subsection{Set-up}

We use the same 10-fold cross-validation event-split in \citet{van-den-berg-markert-2020-context} to facilitate the comparison.

Each fold has 80/10/10 non-overlapping events for train/val/test partition, and sentences from the same event never appear simultaneously in two different subsets within one fold. There are on average 6400/780/790 sentences in train/val/test set respectively. We use 5 different seeds for each method and the F1 score, precision and recall (`biased' is positive class) as the evaluation metrics. For each experiment, a mean value and standard deviation across 5 seeds will be reported if applicable.

Note that in the contrastive learning module, the triplets are constructed only in training set, then the trained model does the sentence-by-sentence inference on the test set (no triplets constructed) to get their sentence embeddings. The CSEs are obtained via inference where no labels are seen during such process. 

We use the same hyper-parameters provided in \citep{van-den-berg-markert-2020-context} to reimplement BERT, RoBERTa and WinSSC baselines. However, for EvCIM, We cut the training epochs from 150 to 75 and increase the batch size from 32 to 64 due to the considerable time usage. For MultiCTX, We trained  use a RoBERTa-based contrastive learning following the implementation in \citep{gao2021simcse}. Due to unavoidable non-deterministic atomic operations in implementation of GAT, the result presented below may cannot be exactly reproduced, but we took an average on our experiments to reflect its range. All models are trained and evaluated on a GeForce GTX 1080 Ti GPU with 11G RAM and Intel(R) Xeon(R) CPU E5-2630 with 128G of RAM.

\subsection{Baselines}

There are few models in sentence-level informational bias detection. \citet{fan-etal-2019-plain} has proposed BASIL dataset and corresponding BERT and RoBERTa benchmarks. \citet{cohan-etal-2019-pretrained} has proposed several models trying to incorporate context in different ways. We will take two of them, WinSSC and their best and also current SOTA model EvCIM, as our baselines. Few other works used BASIL dataset but with objectives other than sentence-level informational detection. Thus we have four baseline models:

\begin{itemize}
    \item \textbf{BERT} \citep{devlin-etal-2019-bert} and \textbf{RoBERTa} \citep{liu2019roberta}: we finetune the individual sentence informational bias detection task on $\text{BERT}_{base}$ and $\text{RoBERTa}_{base}$.
    \item \textbf{WinSSC} \citep{van-den-berg-markert-2020-context}
        
        WinSSC (windowed Sequential Sentence Classification) is a variant of SSC \citep{cohan-etal-2019-pretrained}. We include it as one of the baselines because SSC implements the very natural idea that comes to us when we think of using context: directly inputing sequences of consecutive sentences to BERT. SSC feeds the concatenation of sentences from a chunk of document to pretrained language models (PLMs),and  then classifies each sentence using the embedding of the separator tokens \texttt{[SEP]} at its end. SSC makes non-overlapping chunks while WinSSC makes chunks by overlapping sentences at both ends, which eretains the contextual information for bookended sentences. 
        
    \item \textbf{EvCIM}\label{para:evcim}  : PLM embeddings + BiLSTM
    
        EvCIM (Event Context-Inclusive Model) proposed by \citet{cohan-etal-2019-pretrained} is the SOTA model on BASIL dataset and it also uses the contextual information.
         It takes the average of the last four layers of fine-tuned $\text{RoBERTa}_{base}$ as the sentence embedding, and then uses BiLSTM to encode each article from the same event as the target sentence. Finally it concatenates three article representations and the target sentence embedding to make the sentence-level prediction. Besides using the hyper-parameters from the original paper, we generate the result from a separate set of reasonable hyper parameters. We present below results both from the original paper and from our experiments.

\end{itemize}

\subsection{Our Models}
\begin{itemize}
    \item \textbf{CSE: Contrastive Sentence Embedding}
    
    CSE (Contrastive Sentence Embedding), i.e. sentence embeddings obtained directly from contrastive learning. Here we use to refer to the classification model by a logistic regression on CSE.
    
    \item \textbf{MultiCTX w/o SSGAT}: CSE + BiLSTM
    
    Similar to EvCIM described in Section \ref{para:evcim}, we utilize BiLSTM-encoded context as well as the target sentence to perform the sentence-wise classification. However, instead of the average of the last four layers of fine-tuned $\text{RoBERTa}_{base}$ (RoBERTa embedding, or PLM embedding) in EvCIM, we use CSE in our study. Moreover, we also add news source embeddings before the final fully connected classification layer on top of BiLSTM-encoded in-event article embeddings. 
    
    In the original paper \citep{cohan-etal-2019-pretrained}, adding news source embeddings hurts EvCIM's performance, but because it is useful for EvCIM w/ CSE according to our experiments, we use this version here. This also indicates that CSE has better captured inherent properties of sentences compared to PLM embeddings. CSE can therefore well incorporate extra news media information rather than be disturbed by it.
    
    \item \textbf{MultiCTX w/o CSE}: RoBERTa embeddings + SSGAT
    
    We use the original sentence embedding in EvCIM, which is the
    the average of the last four layers of fine-tuned $\text{RoBERTa}_{base}$ to build the relational sentence graph. We then apply Self-supervised GAT on the graph (SSGAT, Self-supervised Sentence GAT). In other words, we replace CSE in MultiCTX with EvCIM's sentence embedding.
    
    \item \textbf{MultiCTX}: our full model (CSE + SSGAT)
    
    MultiCTX first performs contrastive learning on carefully composed triplets to obtain CSE. It then builds relational sentence graph according to inter-sentence relationships. Finally, MultiCTX applies Self-supervised GAT above to get the final sentence informational bias prediction.

\end{itemize}

\begin{table*}[!htbp]
  \centering
    \begin{tabular}{l|l|p{5em}p{5em}ccc}
    \toprule
     \multicolumn{2}{c}{ \textbf{Model}$^{*}$}  & \textbf{Sentence embedding} & \textbf{Structure to encode context} & \textbf{Precision} & \textbf{Recall} & \textbf{F1} \\
    \midrule
    \multirow{5}{*}{baselines} & $\text{BERT}_{base}$  & NA & No context & $\text{BERT}_{base}$ & $40.44 \pm 1.07^{**}$ & $35.49\pm 0.67$ \\ \cline{2-7}
    &$\text{RoBERTa}_{base}$  & NA & No context & $38.40 \pm 0.64$   & $48.53\pm 1.45$ & $42.13 \pm 1.02$ \\  \cline{2-7}
    & WinSSC & [SEP] embed. & Text chunks & $41.47\pm1.31$& $34.37\pm 0.57$ &$37.58 \pm 0.77$ \\ \cline{2-7}
    
    & {EvCIM (original paper)} 
      & \multirow{2}{*}{RoBERTa}   & \multirow{2}{*}{BiLSTM} &  $39.72\pm 0.59$ &  $49.60 \pm 1.20$ & $44.10\pm 0.15$ \\
      & {EvCIM (our reproduction)} && & $38.40 \pm 0.64$   & $48.53\pm 1.45$  & $42.87 \pm 0.69$ \\ \hline
    \multirow{4}{*}{models}&CSE  & CSE & No context & 47.53 & 40.13 & 43.51  \\ \cline{2-7}
    & MultiCTX w/o SSGAT    &  CSE& BiLSTM &$48.53 \pm 0.73$     & $41.98\pm 0.36$ & $45.01 \pm 0.26$\\ \cline{2-7}
    & MultiCTX w/o CSE &   RoBERTa  & SSGAT &  $46.89 \pm 0.71$  &  $42.88\pm 0.67$  & $44.79 \pm 0.63$ \\ \cline{2-7}
    & MultiCTX   &  CSE  & SSGAT & $47.78 \pm 0.94$   &  $44.50 \pm 0.65$   & $\mathbf{46.08 \pm 0.21}^{***}$ \\
    \bottomrule
    \multicolumn{7}{l}{$^{*}$  All results are implemented or reproduced by ourselves except for EvCIM (original paper)} \\
    \multicolumn{7}{l}{$^{**}$  Mean value and standard deviation across 5 seeds are reported if applicable}  \\
    \multicolumn{7}{l}{$^{***}$  The best result on a single run obtained in our experiments is \textbf{F1=46.74}} \\
    \end{tabular}%
      \caption{Results. It shows the sentence embedding method, the structure to encode the context, the Precision, Recall, and F1 score of each model.}
  \label{tab:res}%
\end{table*}%

The results are shown in Table \ref{tab:res}. By using different sentence embedding methods and the structure to encode contextual information, we are able to demonstrate the respective utility of the two modules, CSE and SSGAT. The table presents the results (Precision, Recall, F1 score) on the four baseline models, our model MultiCTX and its different variants. The four baselines' results are our reproduced results, most of which are close to the original work, except for the EvCIM, so we put both in the table. In addition, we used 5 different random seeds, and the table shows their means and standard deviations (if applicable).

From the results, we can draw the following conclusions:

\paragraph{1. Contrastive learning helps improve sentence embeddings.} 

While keeping the structure to encode context unchanged, models using CSE as the sentence embedding method are the optimal. We see that when the context is not structurally introduced, the performance of CSE (F1=43.51) purely classified by logistic regression is better than that of BERT (F1=35.49) and RoBERTa (F1=42.13); moreover, with BiLSTM as structure to encode the context, MultiCTX w/o SSGAT (F1=45.01) outperforms EvCIM in its original paper \citep{cohan-etal-2019-pretrained}. Since EvCIM uses the mean value of the last four layers of the fine-tuned $\text{RoBERTa}_{base}$ as the sentence embedding, the contrastive learning produces better sentence representations than BERT-based PLMs; the same conclusion can be also drawn from MultiCTX w/o CSE (F1=44.79) vs. MultiCTX (F1=46.08). The reason may be as follows:

\begin{itemize}
    \item BERT-based PLM tends to encode all sentences into a smaller spatial region, which results in a high similarity score for most of the sentence pairs, even for those that are semantically completely unrelated. Specifically, when the sentence embeddings are computed by averaging the word vectors, they are easily dominated by high-frequency words, making it difficult to reflect their original semantics.
    \item Instead of individual sentences, CSE considers for each target sentence a context built up by all its positive and negative counterparts in related triplets. Among them, negative samples provide an article-level context and positive samples provide an event-level context. With the goal of contrastive learning to "distill essence", it learns from its context and naturally suppresses such shallow high-frequency-words features, thus avoiding similar representations of semantically different sentences. 
\end{itemize}

\paragraph{2. Encoding sequential sentences brutally by PLM may fail.}

WinSSC attempts to exploit adjacent sentences by directly feeding sequences of consecutive sentences into the PLM, which is the most failed among all models attempting to incorporate the contextual information. It's even worse (F1=37.58) than the original $\text{RoBERTa}_{base}$ (F1=42.13).There are two possible reasons: First, we take sentence chunks instead of individual sentences as input, and doing so may introduce data reduction. Second, BERT-based pretrained language models are not good at processing long text. They simply join neighboring sentences, which may introduce more noise and complexity rather than help integrate the context. Therefore, brute-force concatenation of sequential sentences can rarely make use of the contextual information, and it probably brings in more noise and reduces the data quantity.

\paragraph{3. Context information can effectively improve performance.}
    
 Except for WinSSC, all other models using some kind of context-introducing structure (BiLSTM or SSGAT) outperform the BERT and RoBERTa baselines. It shows that the introduction of context is indeed helpful for the detection of information bias.

\paragraph{4. SSGAT is the best structure to integrate context.} 

While keeping the sentence embedding method unchanged, the model using SSGAT as the structure to introduce context achieves better results. When both use RoBERTa embeddings, MultiCTX w/o CSE with SSGAT outperforms (F1=44.79) EvCIM with BiLSTM (both F1=44.10 in the original paper and our reproduction F1=42.87); when both use CSE , MultiCTX with SSGAT (F1=46.08) outperforms MultiCTX w/o SSGAT (with BiLSTM, F1=45.01), and CSE without context (F1=43.51). The results prove that our sentence graph structure is more effective in encoding contextual information than sequential models such as BiLSTM.

\paragraph{5. Contrastive learning together with sentence graph achieves the best performance} 

Our full model MultiCTX achieves F1=46.08 in the sentence-level informational bias detection task, significantly outperforms the current State-of-the-Art model EvCIM \citep{cohan-etal-2019-pretrained} (F1=44.10 declared in original paper). Possible reasons are: 1) BiLSTMs are limited to the event context in EvCIM; 2) MultiCTX uses better sentence representations (CSE); 3) MultiCTX incorporates the context in varying degrees explicitly using graph structure and implicitly via contrastive learning.

\section{Ablation Analysis}

We have proved that both CSE and SSGAT are essential for MultiCTX, and in this section, we will further explore roles of different inter-sentence relationships in our model. We keep CSEs fixed and modify our relational sentence graph by removing certain types of edges, and then report the results to see how each part contributes to MultiCTX in our informational bias detection task.

Edge types described in Section \ref{para:edge} can be briefly summarized in two categories: 
\begin{itemize}
    \item Type 1,2 and 3 are discourse relationships
    \item Type 4 is semantic similarity
\end{itemize}  
Besides, they can also be partitioned by level of context:
\begin{itemize}    
    \item Type 3 and 4 are event-level
    \item Type 1 and 2 are neighborhood-level and article-level
\end{itemize}

We will focus on their utility in our ablation study. Table \ref{tab:ablation} shows the ablation results. Horizontally, the first row represents the comparison between discourse relations and semantic similarity, and the second row represents the comparison between event-level context and neighborhood-level context. Vertically, the three ablation analysis experiments in the first column compare the utility of Type 1,2 / Type 3 / Type4 edges. Here we analyze Type 1,2 together.

\begin{table}[htbp]
    \begin{tabular}{rc|p{6em}p{6em}}
    \multirow{1}{*}{\rotatebox[origin=r]{90}{\textcolor{red!60}{\textbf{$\xLeftarrow{\hspace*{5cm}\textbf{Vertical comparison: Edge types}}$}}}}  &   \multicolumn{3}{l}{\textcolor{blue!60}{$\boldsymbol{\xRightarrow{\hspace*{4cm}\textbf{Horizontal comparison}}}$}}\\
      &\textcolor{blue!60}{\textbf{Horizontal}} &  \textcolor{blue!60}{\textbf{Discourse relationship }} & \textcolor{blue!60}{\textbf{Semantic similarity}} 
     \\
   & &  Type [1,2]$^{*}$,3   & Type 4   \\ 
   &\textcolor{red!60}{\textbf{Vertical}} &  \textcolor{red!60}{\textbf{(w/o Type 4)  }} &  \\ \cmidrule[1pt]{2-4}
    &Precision  &  $47.43 \pm 0.96$ &  $47.16 \pm 0.27$  \\
    &Recall     &  $44.39\pm 0.84$  &    $43.47\pm 0.38$ \\
    &F1     &   $45.85\pm0.35$    &  $45.24\pm0.18$ \\
    \cmidrule[1pt]{2-4} 
          & \textcolor{blue!60}{\textbf{Horizontal}} &  \textcolor{blue!60}{\textbf{Event-context}} & \textcolor{blue!60}{\textbf{Neighborhood-context}} \\
       & &  Type 3,4  &  Type [1,2]  \\
      &\textcolor{red!60}{\textbf{Vertical}} &  \textcolor{red!60}{\textbf{(w/o Type 1,2)}} &    \\\cmidrule[1pt]{2-4}
   &Precision &  $47.07\pm 0.99$ &  $47.18 \pm 1.08$  \\
    &Recall     &   $44.64\pm 0.37$  & $44.01\pm 0.91$  \\
    &F1     &   $45.81\pm0.42 $     & $45.53\pm0.29$ \\
    \cmidrule[1pt]{2-4} 
      &  & Type [1,2],4  & \\
      & \textcolor{red!60}{\textbf{Vertical}} &  \textcolor{red!60}{\textbf{(w/o Type 3) }} & \\\cmidrule[1pt]{2-4}
       & Precision & $47.56 \pm 0.62$ &\\
       &Recall & $43.72\pm 0.76$ &\\
    &F1     &    $45.55\pm0.34$ &\\
     \cmidrule[1pt]{2-4}
    & \multicolumn{3}{p{18em}}{$^{*}$  Type 1,2 represent neighborhood-context, so we treat them as a whole.}\\
    \end{tabular}%
    \caption{Ablation study on different types of edges in SSGAT. Horizontally, the first two rows compare discourse relation vs. semantic similarity and event-context vs. neighborhood-context respectively. Vertically, the first column compares the utility of the Type[1,2] vs. Type 3 vs. Type 4 edges. Mean and standard deviation across 5 seeds are reported.}
  \label{tab:ablation}%
\end{table}%

In addition to the numerical results, we also want to better analyze the utility of various types of edges through the perspective of plotting, so we take an event in the dataset and connect different types of edges between its sentence nodes And draw the corresponding picture for comparison. Figure \ref{tab:ablationfig} shows these graphs, where purple nodes are unbiased sentences and yellow nodes are biased sentences. The nodes of each graph are the same, the only difference is the edge types.

\begin{table}[htbp]
    \addtolength{\leftskip} {-0.7cm}
    \begin{tabular}{rc|p{8em}p{8em}}
    \multirow{1}{*}{\rotatebox[origin=r]{90}{\textcolor{red!60}{\textbf{$\xLeftarrow{\hspace*{12cm}\textbf{Vertical comparison: Edge types}}$}}}}  &   \multicolumn{3}{l}{\textcolor{blue!60}{$\boldsymbol{\xRightarrow{\hspace*{5cm}\textbf{Horizontal comparison}}}$}}\\
      &\textcolor{blue!60}{\textbf{Horizontal }} &  \textcolor{blue!60}{\textbf{Discourse relationship }} & \textcolor{blue!60}{\textbf{Semantic similarity}} 
     \\
    & &  Type [1,2]$^{*}$,3   & Type 4   \\ 
   &\textcolor{red!60}{\textbf{Vertical}} &  \textcolor{red!60}{\textbf{(w/o Type 4)  }} &  \\ \cmidrule[1pt]{2-4}
   &&\begin{minipage}{\linewidth}
      \includegraphics[width=\linewidth]{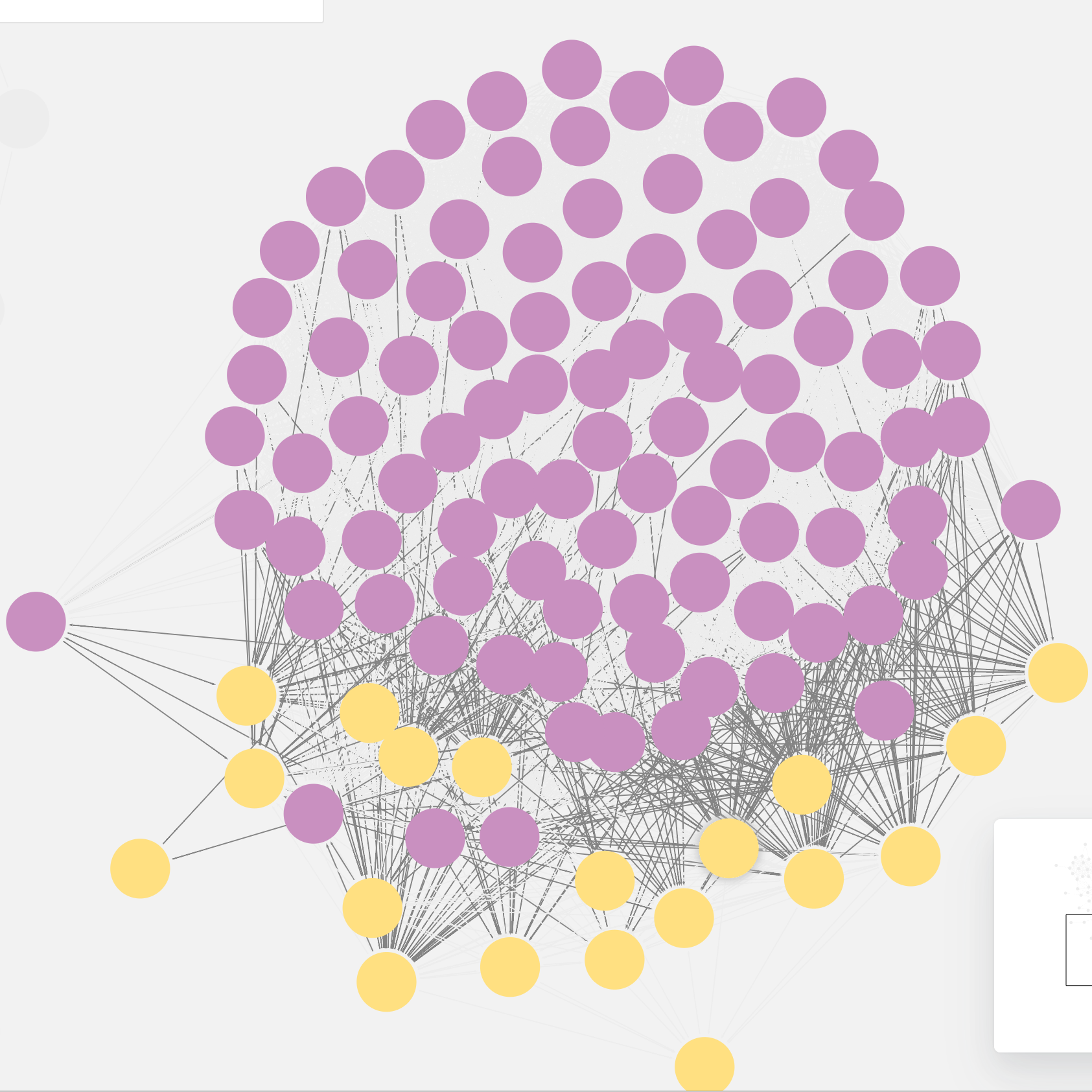}\captionof{figure}{}\label{subfig:ab123}
    \end{minipage} &\begin{minipage}{\linewidth}
      \includegraphics[width=\linewidth]{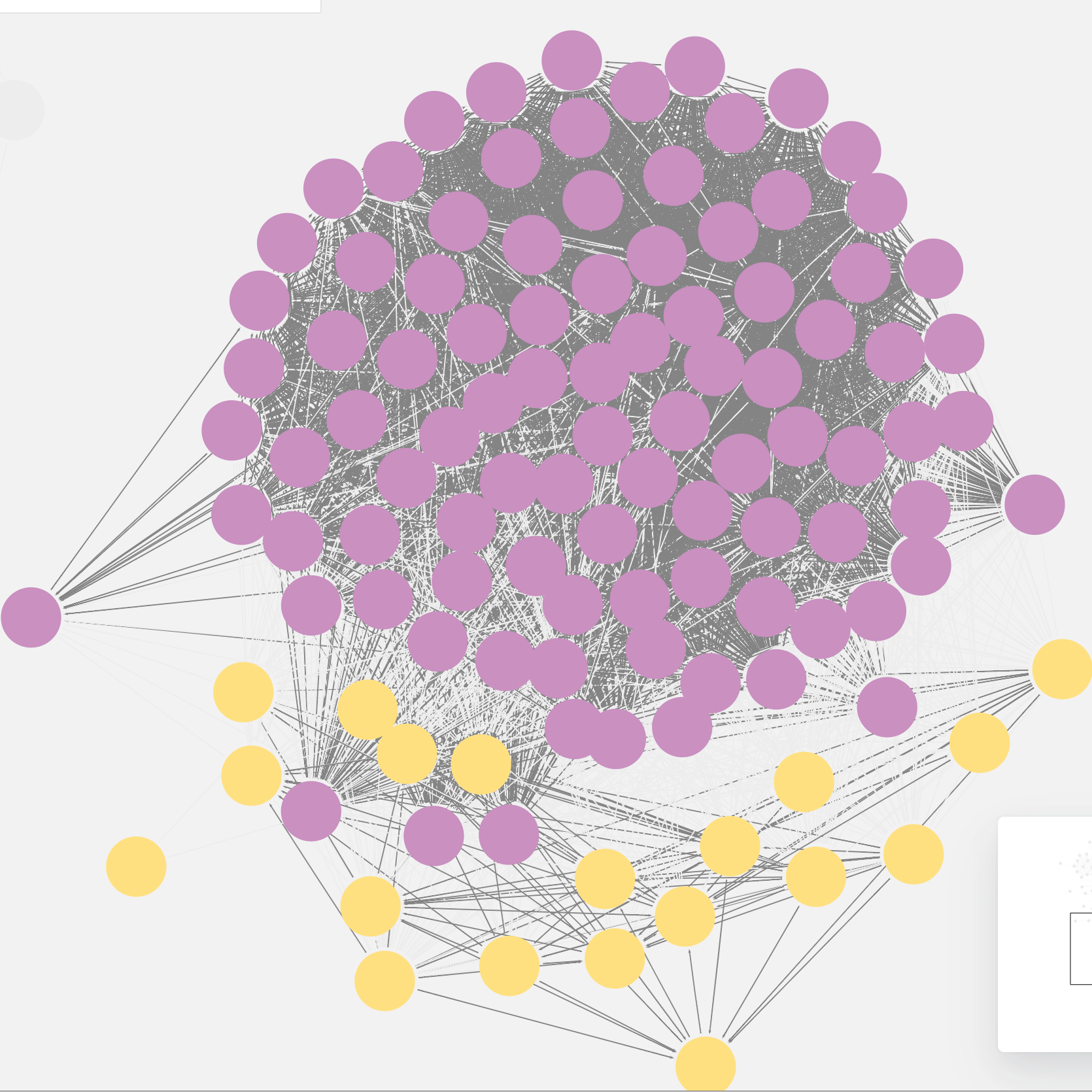}\captionof{figure}{}\label{subfig:ab4}
    \end{minipage} \\
    \cmidrule[1pt]{2-4} 
          & \textcolor{blue!60}{\textbf{Horizontal }} &  \textcolor{blue!60}{\textbf{Event-context}} & \textcolor{blue!60}{\textbf{Neighborhood-context}} \\
       & &  Type 3,4  &  Type [1,2]  \\
      &\textcolor{red!60}{\textbf{Vertical}} &  \textcolor{red!60}{\textbf{(w/o Type 1,2)}} &    \\\cmidrule[1pt]{2-4}
   &&\begin{minipage}{\linewidth}
      \includegraphics[width=\linewidth]{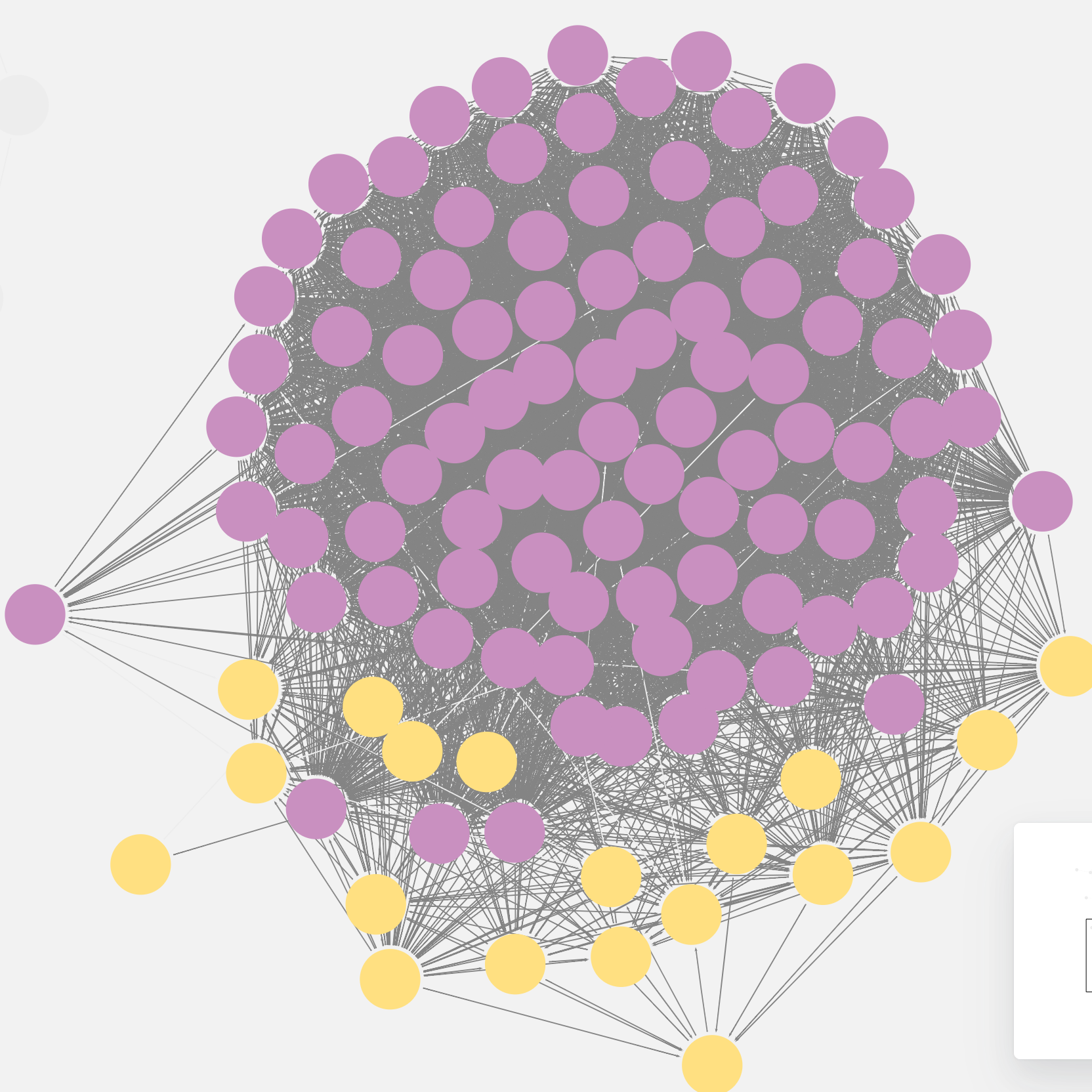}\captionof{figure}{}\label{subfig:ab34}
    \end{minipage} &\begin{minipage}{\linewidth}
      \includegraphics[width=\linewidth]{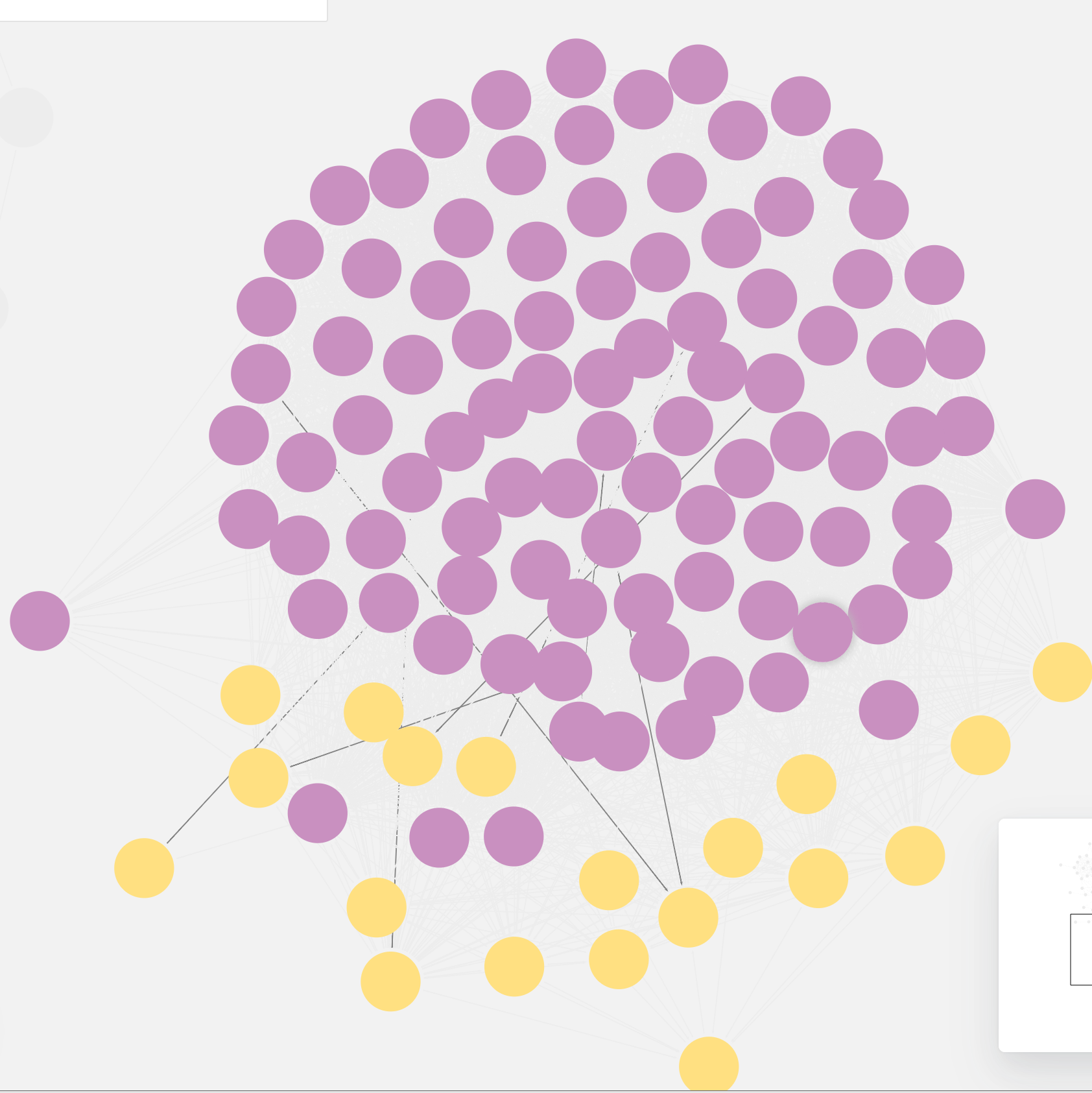}\captionof{figure}{}\label{subfig:ab12}
    \end{minipage} \\
    \cmidrule[1pt]{2-4} 
      &  & Type [1,2],4  & Type 3\\
      & \textcolor{red!60}{\textbf{Vertical}} &  \textcolor{red!60}{\textbf{(w/o Type 3) }} & \textcolor{red!60}{\textbf{only Type 3 }}\\\cmidrule[1pt]{2-4}
   &&\begin{minipage}{\linewidth}
      \includegraphics[width=\linewidth]{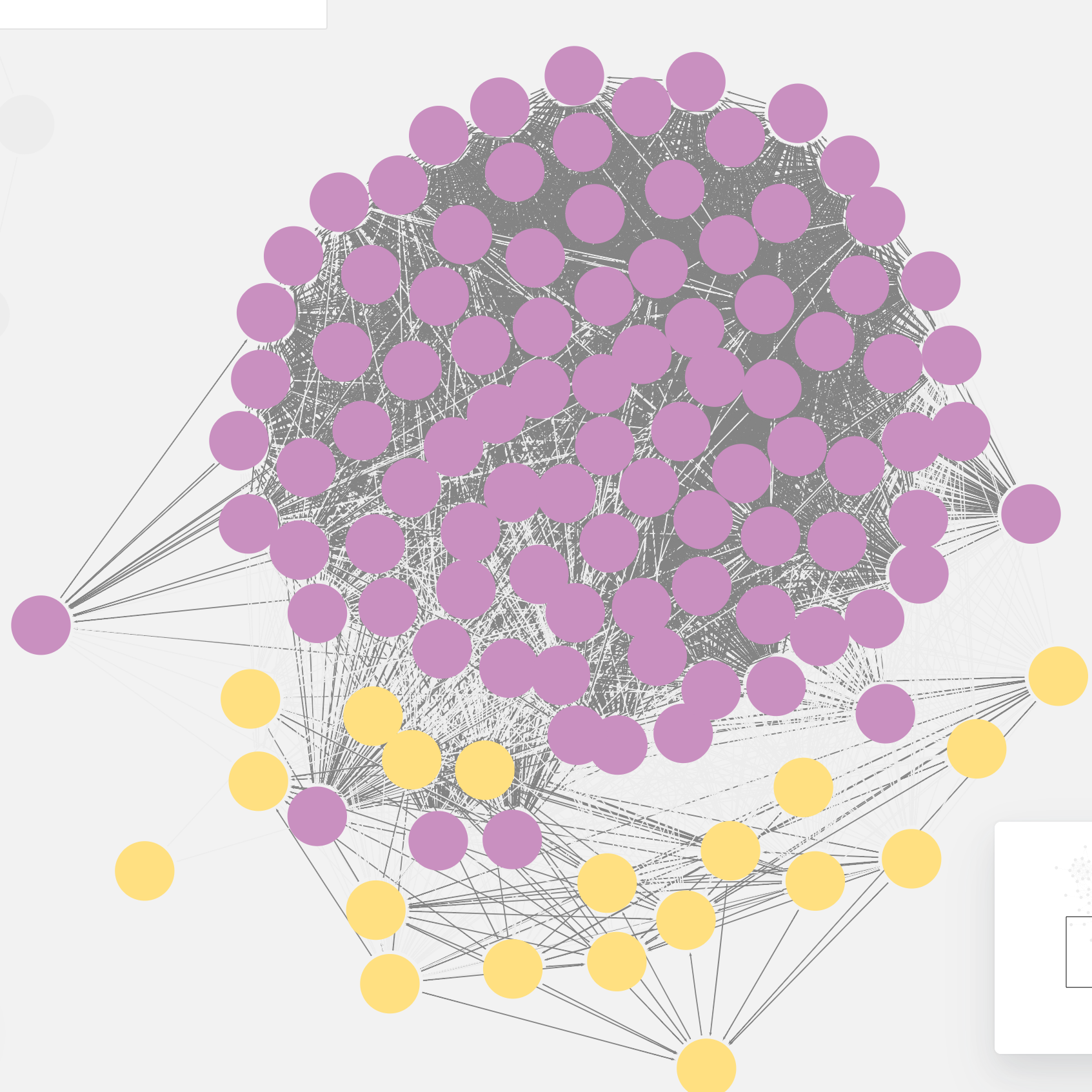}\captionof{figure}{}\label{subfig:ab124}
    \end{minipage} &\begin{minipage}{\linewidth}
      \includegraphics[width=\linewidth]{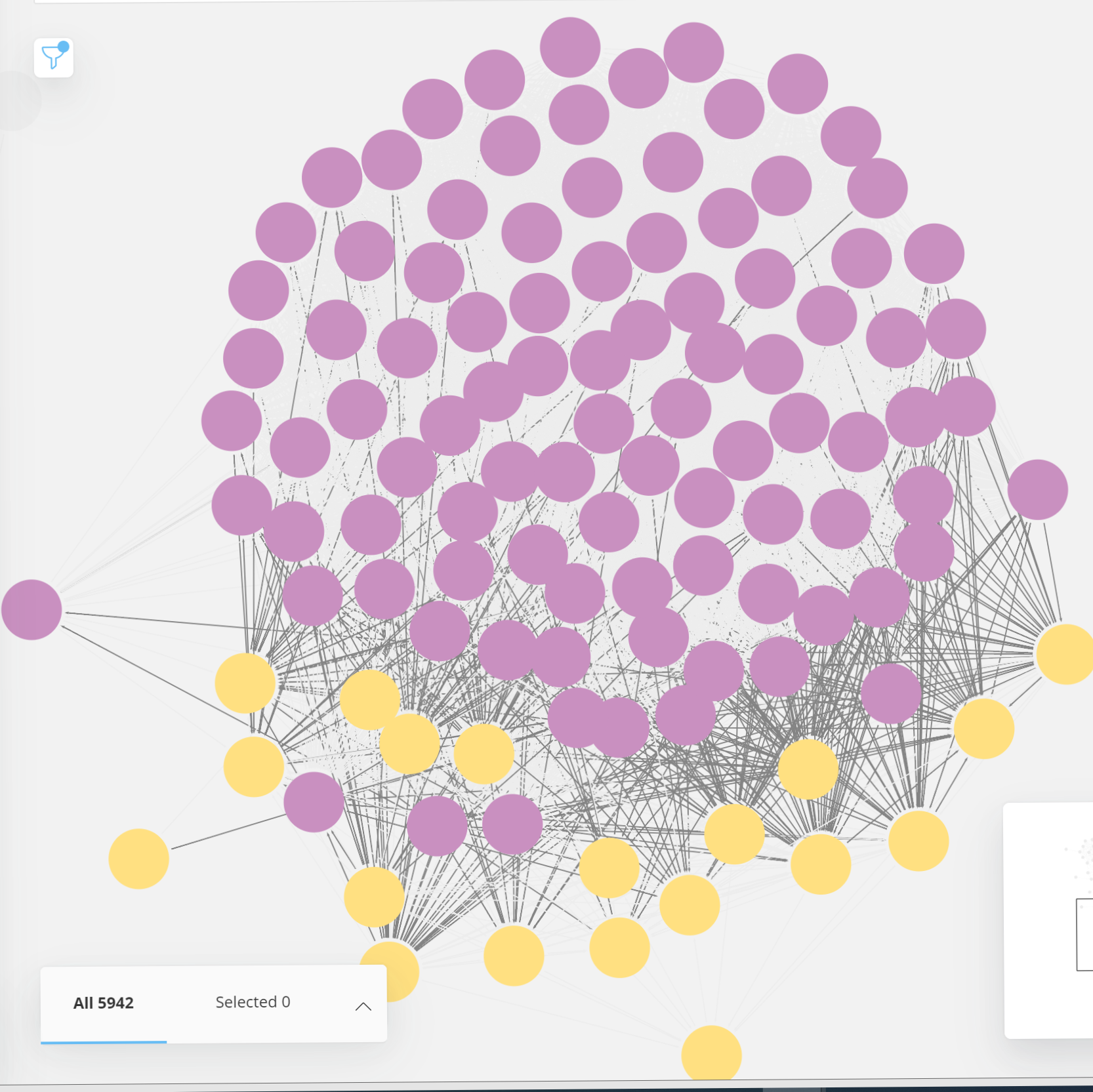}\captionof{figure}{}\label{subfig:ab3}
    \end{minipage} \\
     \cmidrule[1pt]{2-4}
    \end{tabular}%
    \caption{Graphs of ablation studies on different types of edges in SSGAT. They are based on one same event in the dataset. Purple nodes are unbiased sentences and yellow nodes are biased sentences. the nodes of each graph are the same, the only difference is the edge types. }
  \label{tab:ablationfig}%
\end{table}%

We can conclude that, 

\begin{itemize}

    \item \textbf{Discourse relationship contributes more than semantic similarity to SSGAT.}
    
    SSGAT with only discourse relationships (F1=45.85) still has close performance to the full MultiCTX, while SSGAT with only semantic similarity edges (F1=45.25) suffers a considerable decrease in its performance. Note that the semantic similarity is calculated based on CSEs, so SSGAT with only such edges didn't add much extra information but may introduce duplication. It can be explained by Figure \ref{subfig:ab4}: connections are mostly within non-biased nodes while inter-communication between biased/non biased nodes are more frequent in Figure \ref{subfig:ab123}.

    \item \textbf{Event-level context is more important than neighborhood-level context.} 
    
    While they are both important according to our results, global event-level context contributes more than local neighborhood-level context. SSGAT with only adjacent sentences (Type 1,2) obtains F1=45.53 and with only Type 3,4 gets F1=45.81. The result is intuitive because edges of type 3,4 not only include adjacent sentences within article, but also extend to the whole event. We can also see the rare presence of edges of Type 1,2 in Figure \ref{subfig:ab12} compared with the closely linked graph in Figure  \ref{subfig:ab34}.

    \item \textbf{Entity continuation is the most important edge type.}
    
    Among three ablation experiments removing respectively edges of Type 4 (F1=45.85), Type 1,2 (F1=45.81) and Type 3 (F1=45.55), the last one without Type 3 (entity continuation) suffers the largest performance drop. It suggests that entity continuation, or, coreference is the most important relation in our setting.
    We can clearly see that Type 3 edges are the main reason for inter-class communication from Figure \ref{subfig:ab3} and Figure \ref{subfig:ab124}.

\end{itemize}

\section{Related Work}

\paragraph{Media bias Detection.}

With the rise of deep learning, neural-based aproaches are broadly used in media bias detection. \citet{iyyer-etal-2014-political} used RNNs to aggregate
the polarity of each word to predict political ideology on sentence-level. \citet{gangula-etal-2019-detecting}
made use of headline attention to classify article
bias. \citet{li-goldwasser-2019-encoding} captured social information by Graph Convolutional Network to identify political bias in news articles. \citet{fan-etal-2019-plain} used BERT and RoBERTa and \citet{van-den-berg-markert-2020-context} used BiLSTMs as well as BERT-based models to detect sentence-level informational bias.
\paragraph{Contextual information in media bias detection.} 
Contextual information is explored, though primarily, in media bias detection. \citet{baly2020detect} employed an adversarial news media adaptation using triplet loss; \citet{kulkarni-etal-2018-multi} proposed an attention based model to capture views from news articles' title, content and link structure; \citet{chen-etal-2020-detecting} explored the impact of sentence-level bias to article-level bias; \citet{li-goldwasser-2019-encoding} encoded social information using GCN; \citet{baly-etal-2018-predicting} made use of news media's cyber-features in news factuality prediction; \citet{10.1145/3366423.3380158} explored cross-media context by a news article graph.

Sentence-level informational bias is under-studied by only a few research and the methods described above are not applicable on this task. In order to infuse contextual information, we refer to extractive summarization \citet{10.1145/3397271.3401327} and \citet{christensen-etal-2013-towards} which used sentence graph to encode context.

\section{Conclusion}

Our work focus on incorporating different levels of context: neighborhood-level, article-level and event-level in sentence-level informational bias detection. We proposed MultiCTX, a model composed of contrastive learning and relational sentence graph attention network to encode such multi-level context at different stages. 

Our model (F1=46.08) significantly outperforms the current state-of-the-art model (F1=44.10) by 2 percentage points. Therefore, we conclude that our model successfully learns from contextual information and that multi-level contextual information can effectively improves 
the identification of sentence-level informational bias. 

Moreover, our design aims at simulate the way people learn new things in real life: learn from multiple news reports covering the whole event to form a general picture and then use the past experience to reason about the unknown.

\bibliography{aaai22.bib}

\begin{thebibliography}{19}
\providecommand{\natexlab}[1]{#1}

\bibitem[{Baly et~al.(2020{\natexlab{a}})Baly, Da~San~Martino, Glass, and
  Nakov}]{baly-etal-2020-detect}
Baly, R.; Da~San~Martino, G.; Glass, J.; and Nakov, P. 2020{\natexlab{a}}.
\newblock We Can Detect Your Bias: Predicting the Political Ideology of News
  Articles.
\newblock In \emph{Proceedings of the 2020 Conference on Empirical Methods in
  Natural Language Processing (EMNLP)}, 4982--4991. Online: Association for
  Computational Linguistics.

\bibitem[{Baly et~al.(2018)Baly, Karadzhov, Alexandrov, Glass, and
  Nakov}]{baly-etal-2018-predicting}
Baly, R.; Karadzhov, G.; Alexandrov, D.; Glass, J.; and Nakov, P. 2018.
\newblock Predicting Factuality of Reporting and Bias of News Media Sources.
\newblock In \emph{Proceedings of the 2018 Conference on Empirical Methods in
  Natural Language Processing}, 3528--3539. Brussels, Belgium: Association for
  Computational Linguistics.

\bibitem[{Baly et~al.(2020{\natexlab{b}})Baly, Martino, Glass, and
  Nakov}]{baly2020detect}
Baly, R.; Martino, G. D.~S.; Glass, J.; and Nakov, P. 2020{\natexlab{b}}.
\newblock We Can Detect Your Bias: Predicting the Political Ideology of News
  Articles.
\newblock arXiv:2010.05338.

\bibitem[{Chen et~al.(2020)Chen, Al~Khatib, Stein, and
  Wachsmuth}]{chen-etal-2020-detecting}
Chen, W.-F.; Al~Khatib, K.; Stein, B.; and Wachsmuth, H. 2020.
\newblock Detecting Media Bias in News Articles using {G}aussian Bias
  Distributions.
\newblock In \emph{Findings of the Association for Computational Linguistics:
  EMNLP 2020}, 4290--4300. Online: Association for Computational Linguistics.

\bibitem[{Christensen et~al.(2013)Christensen, {Mausam}, Soderland, and
  Etzioni}]{christensen-etal-2013-towards}
Christensen, J.; {Mausam}; Soderland, S.; and Etzioni, O. 2013.
\newblock Towards Coherent Multi-Document Summarization.
\newblock In \emph{Proceedings of the 2013 Conference of the North {A}merican
  Chapter of the Association for Computational Linguistics: Human Language
  Technologies}, 1163--1173. Atlanta, Georgia: Association for Computational
  Linguistics.

\bibitem[{Cohan et~al.(2019)Cohan, Beltagy, King, Dalvi, and
  Weld}]{cohan-etal-2019-pretrained}
Cohan, A.; Beltagy, I.; King, D.; Dalvi, B.; and Weld, D. 2019.
\newblock Pretrained Language Models for Sequential Sentence Classification.
\newblock In \emph{Proceedings of the 2019 Conference on Empirical Methods in
  Natural Language Processing and the 9th International Joint Conference on
  Natural Language Processing (EMNLP-IJCNLP)}, 3693--3699. Hong Kong, China:
  Association for Computational Linguistics.

\bibitem[{Devlin et~al.(2019)Devlin, Chang, Lee, and
  Toutanova}]{devlin-etal-2019-bert}
Devlin, J.; Chang, M.-W.; Lee, K.; and Toutanova, K. 2019.
\newblock {BERT}: Pre-training of Deep Bidirectional Transformers for Language
  Understanding.
\newblock In \emph{Proceedings of the 2019 Conference of the North {A}merican
  Chapter of the Association for Computational Linguistics: Human Language
  Technologies, Volume 1 (Long and Short Papers)}, 4171--4186. Minneapolis,
  Minnesota: Association for Computational Linguistics.

\bibitem[{Fan et~al.(2019)Fan, White, Sharma, Su, Choubey, Huang, and
  Wang}]{fan-etal-2019-plain}
Fan, L.; White, M.; Sharma, E.; Su, R.; Choubey, P.~K.; Huang, R.; and Wang, L.
  2019.
\newblock In Plain Sight: Media Bias Through the Lens of Factual Reporting.
\newblock In \emph{Proceedings of the 2019 Conference on Empirical Methods in
  Natural Language Processing and the 9th International Joint Conference on
  Natural Language Processing (EMNLP-IJCNLP)}, 6343--6349. Hong Kong, China:
  Association for Computational Linguistics.

\bibitem[{Gangula, Duggenpudi, and Mamidi(2019)}]{gangula-etal-2019-detecting}
Gangula, R. R.~R.; Duggenpudi, S.~R.; and Mamidi, R. 2019.
\newblock Detecting Political Bias in News Articles Using Headline Attention.
\newblock In \emph{Proceedings of the 2019 ACL Workshop BlackboxNLP: Analyzing
  and Interpreting Neural Networks for NLP}, 77--84. Florence, Italy:
  Association for Computational Linguistics.

\bibitem[{Gao, Yao, and Chen(2021)}]{gao2021simcse}
Gao, T.; Yao, X.; and Chen, D. 2021.
\newblock SimCSE: Simple Contrastive Learning of Sentence Embeddings.
\newblock arXiv:2104.08821.

\bibitem[{Huang et~al.(2020)Huang, Liu, Lee, Calderon~Alvarado, and
  Chen}]{10.1145/3366423.3380158}
Huang, Y.-H.; Liu, T.-W.; Lee, S.-R.; Calderon~Alvarado, F.~H.; and Chen, Y.-S.
  2020.
\newblock Conquering Cross-Source Failure for News Credibility: Learning
  Generalizable Representations beyond Content Embedding.
\newblock In \emph{Proceedings of The Web Conference 2020}, WWW '20, 774–784.
  New York, NY, USA: Association for Computing Machinery.
\newblock ISBN 9781450370233.

\bibitem[{Iyyer et~al.(2014)Iyyer, Enns, Boyd-Graber, and
  Resnik}]{iyyer-etal-2014-political}
Iyyer, M.; Enns, P.; Boyd-Graber, J.; and Resnik, P. 2014.
\newblock Political Ideology Detection Using Recursive Neural Networks.
\newblock In \emph{Proceedings of the 52nd Annual Meeting of the Association
  for Computational Linguistics (Volume 1: Long Papers)}, 1113--1122.
  Baltimore, Maryland: Association for Computational Linguistics.

\bibitem[{Kim and Oh(2021)}]{kim2021how}
Kim, D.; and Oh, A. 2021.
\newblock How to Find Your Friendly Neighborhood: Graph Attention Design with
  Self-Supervision.
\newblock In \emph{International Conference on Learning Representations}.

\bibitem[{Kulkarni et~al.(2018)Kulkarni, Ye, Skiena, and
  Wang}]{kulkarni-etal-2018-multi}
Kulkarni, V.; Ye, J.; Skiena, S.; and Wang, W.~Y. 2018.
\newblock Multi-view Models for Political Ideology Detection of News Articles.
\newblock In \emph{Proceedings of the 2018 Conference on Empirical Methods in
  Natural Language Processing}, 3518--3527. Brussels, Belgium: Association for
  Computational Linguistics.

\bibitem[{Li and Goldwasser(2019)}]{li-goldwasser-2019-encoding}
Li, C.; and Goldwasser, D. 2019.
\newblock Encoding Social Information with Graph Convolutional Networks
  for{P}olitical Perspective Detection in News Media.
\newblock In \emph{Proceedings of the 57th Annual Meeting of the Association
  for Computational Linguistics}, 2594--2604. Florence, Italy: Association for
  Computational Linguistics.

\bibitem[{Liu et~al.(2019)Liu, Ott, Goyal, Du, Joshi, Chen, Levy, Lewis,
  Zettlemoyer, and Stoyanov}]{liu2019roberta}
Liu, Y.; Ott, M.; Goyal, N.; Du, J.; Joshi, M.; Chen, D.; Levy, O.; Lewis, M.;
  Zettlemoyer, L.; and Stoyanov, V. 2019.
\newblock RoBERTa: A Robustly Optimized BERT Pretraining Approach.
\newblock arXiv:1907.11692.

\bibitem[{van~den Berg and Markert(2020)}]{van-den-berg-markert-2020-context}
van~den Berg, E.; and Markert, K. 2020.
\newblock Context in Informational Bias Detection.
\newblock In \emph{Proceedings of the 28th International Conference on
  Computational Linguistics}, 6315--6326. Barcelona, Spain (Online):
  International Committee on Computational Linguistics.

\bibitem[{Zhao et~al.(2020{\natexlab{a}})Zhao, Liu, Gao, Jin, Du, Zhao, Zhang,
  and Haffari}]{summpip}
Zhao, J.; Liu, M.; Gao, L.; Jin, Y.; Du, L.; Zhao, H.; Zhang, H.; and Haffari,
  G. 2020{\natexlab{a}}.
\newblock \emph{SummPip: Unsupervised Multi-Document Summarization with
  Sentence Graph Compression}, 1949–1952.
\newblock New York, NY, USA: Association for Computing Machinery.
\newblock ISBN 9781450380164.

\bibitem[{Zhao et~al.(2020{\natexlab{b}})Zhao, Liu, Gao, Jin, Du, Zhao, Zhang,
  and Haffari}]{10.1145/3397271.3401327}
Zhao, J.; Liu, M.; Gao, L.; Jin, Y.; Du, L.; Zhao, H.; Zhang, H.; and Haffari,
  G. 2020{\natexlab{b}}.
\newblock \emph{SummPip: Unsupervised Multi-Document Summarization with
  Sentence Graph Compression}, 1949–1952.
\newblock New York, NY, USA: Association for Computing Machinery.
\newblock ISBN 9781450380164.

\end{thebibliography}

\end{document}